\documentclass{article}

\usepackage{PRIMEarxiv}

\usepackage[utf8]{inputenc} % allow utf-8 input
\usepackage[T1]{fontenc}    % use 8-bit T1 fonts
\usepackage{hyperref}       % hyperlinks
\usepackage{url}            % simple URL typesetting
\usepackage{booktabs}       % professional-quality tables
\usepackage{amsfonts}       % blackboard math symbols
\usepackage{nicefrac}       % compact symbols for 1/2, etc.
\usepackage{microtype}      % microtypography
\usepackage{fancyhdr}       % header
\usepackage{graphicx}       % graphics
\graphicspath{{media/}}     % organize your images and other figures under media/ folder
\usepackage[authoryear]{natbib}
\usepackage{amssymb}
\usepackage{amsmath}
\usepackage{subcaption}
\usepackage{todonotes}
\usepackage{booktabs}
\usepackage{tabularx}
\usepackage{multirow}
\usepackage{siunitx}
\usepackage{url}
\title{Natural Language Counterfactual Explanations for Graphs Using Large Language Models}

\author{
  Flavio Giorgi\\
  Sapienza University of Rome\\
  Rome, Italy\\
  \texttt{giorgi@di.uniroma1.it} \\
   \And
  Cesare Campagnano \\
   Sapienza University of Rome\\
  Rome, Italy \\
  \texttt{campagnano@di.uniroma1.it} \\
     \And
  Fabrizio Silvestri \\
   Sapienza University of Rome\\
  Rome, Italy \\
  \texttt{fsilvestri@diag.uniroma1.it} \\
     \And
  Gabriele Tolomei \\
   Sapienza University of Rome\\
  Rome, Italy \\
  \texttt{toloemi@di.uniroma1.it} \\
}

\newcommand{\params}{\boldsymbol{\theta}}

\newcommand{\model}{f_{\params}}

\begin{document}
\maketitle

\begin{abstract}
Explainable Artificial Intelligence (XAI) has emerged as a critical area of research to unravel the opaque inner logic of (deep) machine learning models. 
Among the various XAI techniques proposed in the literature, counterfactual explanations stand out as one of the most promising approaches. However, these ``what-if'' explanations are frequently complex and technical, making them difficult for non-experts to understand and, more broadly, challenging for humans to interpret.
To bridge this gap, in this work, we exploit the power of open-source Large Language Models to generate natural language explanations when prompted with valid counterfactual instances produced by state-of-the-art explainers for graph-based models. 
Experiments across several graph datasets and counterfactual explainers show that our approach effectively produces accurate natural language representations of counterfactual instances, as demonstrated by key performance metrics.
\end{abstract}

% keywords can be removed

\section{Introduction}
\label{sec:intro}

In recent years, Machine Learning have become deeply embedded in various facets of our daily lives, influencing decisions ranging from personalized recommendations to critical judgments in healthcare and finance. This pervasive integration has raised significant concerns regarding transparency and accountability, leading to legislative actions such as the European Union's General Data Protection Regulation (GDPR) \citep{voigt2017eu}, which emphasizes the right to explanation for automated decision-making processes. Similarly, the upcoming EU Artificial Intelligence Act aims to regulate AI systems, mandating explainability and interpretability in high-risk applications \citep{eu2021ai}.

In the realm of eXplainable Artificial Intelligence (XAI), numerous techniques have been proposed to address this need. Among them, \textit{counterfactual explanations} \citep{wachter2017hjlt} stand out as one of the most promising \textit{post-hoc} methods. 
The core idea is to explain a model's prediction by identifying a \textit{counterfactual example} -- i.e., the minimal change to the input that leads to a different prediction. 
Counterfactual explanations have been successfully applied to unveil the inner workings of predictive models having various degrees of complexity, ranging from ensembles of decision trees \citep{tolomei2017kdd} and multi-layer perceptrons \citep{montavon2018methods} to more sophisticated models like transformers \citep{chefer2020transformer}.

Despite these advancements, a significant challenge remains: translating algorithmically generated explanations into a ``format'' that is accessible and comprehensible to end-users who may not possess technical expertise. 
\cite{fredes2024using} takes a first step in this direction by leveraging the generative capabilities of Large Language Models (LLMs) to produce user-friendly counterfactual explanations in natural language. 

Building on the work above, we tackle the more ambitious challenge of generating natural language explanations from counterfactual examples tailored for graph neural networks (GNNs). 
Indeed, counterfactual examples derived from graph inputs are inherently more complex than those from tabular data, given the intricate relationships and structures they capture (e.g., nodes, edges, and their dependencies). This complexity presents unique challenges that demand a specialized approach to translate them into universally comprehensible natural language explanations.

Moreover, graph data is prevalent across various domains such as financial decision-making \citep{cao2020spectral}, fraud detection \citep{wang2021review}, and biology \citep{fout2017protein}, where GNN-based models have shown remarkable predictive performance by encoding high-order structural relationships within their learned graph representations.

Specifically, we consider the counterfactual examples output by a generic graph counterfactual explainer designed for node classification tasks using GNNs.
Then, we instruct several open-source LLMs to translate these ``raw'' counterfactual examples into coherent natural language explanations that are accessible also to non-expert users. 
To evaluate the quality of the generated explanations, we introduce a set of novel metrics that measure how accurately our method maps the counterfactual examples to their corresponding natural language descriptions. As openly generated text might be hard to evaluate using automatic evaluation metrics \citep{chiang-lee-2023-large}, we also validate this evaluation through human judgment.
Extensive experiments conducted using two graph counterfactual explainers for node classification -- CF-GNNExplainer \citep{lucic2022cf} and its extension for node features, CF-GNNFeatures \citep{giorgi2024} -- across several graph datasets and multiple open-source LLMs demonstrate that our method can effectively support decision-making processes in critical domains through the generation of natural language explanations.

To summarize, our main contributions are as follows: \begin{itemize} 
    \item We present a method for translating counterfactual explanations for graphs into natural language using state-of-the-art open-source LLMs. This is, to our knowledge, the first work proposing the use of LLMs as means for converting the output of a GNN model to a human-readable format. 
    \item We define novel metrics to properly assess the effectiveness of these explanations. 
    \item We perform an extensive evaluation of our approach, which includes varying the sizes of LLMs, datasets, and explanation methods.
    
\end{itemize}

This paper is structured as follows: in Section~\ref{sec:related}, we summarize related work; In Section~\ref{sec:method}, we describe our method, which is validated through extensive experiments in Section~\ref{sec:experiments}. We discuss the practical implications of our method in Section~\ref{sec:discussion}. Finally, Section~\ref{sec:conclusion} concludes our work and proposes future directions.

The code to reproduce our experiments is available at \href{https://github.com/flaat/llm-graph-cf}{https://github.com/flaat/llm-graph-cf}.

\section{Related Work}\label{sec:related}
Counterfactual explainability has emerged as a pivotal approach for interpreting complex machine learning models by illustrating how changes in input variables can lead to different outcomes. Despite this, a significant challenge persists: making these explanations accessible and understandable to a broad audience, particularly non-technical users who may struggle with abstract mathematical concepts and technical jargon. The intricacy of counterfactual explanations often hinders their comprehension among laypersons, limiting their practical utility in real-world applications where user trust and transparency are paramount.

The advent of large language models (LLMs), such as GPT-4 and its successors, has revolutionized the field of natural language processing. These models possess an unparalleled capacity for understanding and generating human-like text, making them invaluable tools for translating complex technical information into plain language. Their widespread availability and ease of integration into various platforms further enhance their appeal for tasks requiring sophisticated language generation and interpretation.

Leveraging the immense capabilities of LLMs presents a promising solution to the accessibility problem in counterfactual explainability. By converting intricate data-driven explanations into natural language narratives, LLMs can make the insights derived from machine learning models more digestible for non-technical users. This translation not only aids in user comprehension but also fosters greater trust in automated systems by promoting transparency.

In this context, the pioneering work done by \cite{fredes2024using} marks a significant milestone. Their research laid the foundations for utilizing LLMs to transform data-based counterfactual explanations into coherent, user-friendly language. After generating the counterfactual examples, the researchers aimed to identify the primary causal factors deduced from these examples that led to the user's differing classification. To achieve this, they input both the set of counterfactual examples and the original user data into a Large Language Model (LLM), instructing it to produce a list of the main reasons why the user was classified differently. Once the LLM generated this list, they meticulously verified its accuracy and identified the most relevant causes contributing to the explanation.

Subsequently, they synthesized all the information produced and leveraged the LLM once more to generate a final explanation articulated in plain language. This explanation was crafted to emphasize actionable steps that the user could take to alter their input data or behavior, thereby changing their classification to the desired category. By doing so, they not only provided a transparent rationale behind the classification but also offered practical guidance for the user to achieve a favorable outcome. 
However, their work focuses solely on the relatively straightforward case of translating counterfactual examples derived from a single, well-known tabular dataset. Furthermore, they used a proprietary LLM model (GPT-4o) making it hard to replicate their approach.

To the best of the authors' knowledge, there are no papers in the State-of-the-Art that address the problem of translating the explanations generated via counterfactual explainers into natural language explanations; for this reason, we firmly believe that our work can be useful to the community.

\section{From Counterfactual Examples to Natural Language Explanations} \label{sec:method}
\subsection{The Counterfactual Explanation Problem}
The counterfactual explanation problem in a classification task is described as follows. Given a sample $x$ and a predictive model $\model$ parametrized by $\params$ -- hereinafter referred to as \textit{oracle} -- the goal is to find a sample $x'\neq x$ such that $\model(x) \neq \model(x')$. This $x'$ is called a \textit{counterfactual example} for $x$. 
Among all potential counterfactual examples (if any), we assume the existence of a counterfactual example generator $g$, which takes as input the original instance $x$ and returns a counterfactual $x^*$, where the distance $d(x, x^*)$ is minimized, or $\perp$ if no valid counterfactual example exists. 
The distance function $d(\cdot, \cdot)$ ensures that the counterfactual sample $x^*$ remains as close as possible to the original factual sample $x$.

Note that, in this work, we focus on graph inputs. Specifically, each sample $x$ and its counterfactual(s) $x'$ can be represented as $G(V, E)$ and $G'(V', E')$, respectively. Therefore, a counterfactual example for an input graph $G$ will be a new graph $G'$, where either the node features, the structural links, or both differ from the original, while still satisfying the counterfactual criterion of altering the model's prediction. This introduces additional challenges as the modifications can affect both the node-level properties and the overall graph topology.

\subsection{Proposed Method}

We assume to have an oracle $\model$ for a node classification task, specifically a graph neural network trained on a given input graph. 
For any node instance $x$, we know both its predicted label $\hat{y}$, such that $\model(x) = \hat{y}$, and its optimal counterfactual example $x^* = g(x)$, which is generated by the counterfactual explainer $g$.

To generate the natural language explanation ($e(x^*)$) associated with the generic counterfactual example ($x^*$), we feed a pre-trained LLM $m$ with the factual ($x$) and counterfactual ($x^*$) instances along with a specific prompt $p$, i.e., $e(x^*) = m(p, x, x^*)$. 

One of the critical challenges of this approach is how to validate the quality of the generated natural language explanations through the LLM. 
In the following section, we introduce the new evaluation framework proposed in this work.

\subsection{A New Evaluation Framework}
Given the novelty of the problem we are tackling, to the best of our knowledge, no established quality metrics exist in the literature to rigorously evaluate explanations generated by LLMs for graph counterfactuals. 
The only commonly adopted approach is based on subjective human judgment, which may introduce variability and bias in the assessment process. To address this gap and ensure a more systematic evaluation, we have developed a suite of novel metrics that provide a quantitative and objective assessment of the language model’s capability to articulate pairs of factual and counterfactual graphs into coherent and informative natural language explanations.

For clarity, we denote the factual graph as $G$ and the corresponding counterfactual graph as $G'$. To compute these metrics effectively, we structured our prompts to include a request for the language model to populate a predefined dictionary with essential graph information. This dictionary encompasses the target node of the classification task, its original class in the factual graph, its modified class in the counterfactual graph, the neighborhood of the target node in both $G$ and $G'$, and the set of features associated with the target node in each scenario.

By doing so, we aim to evaluate the language model’s ability to discern and convey the critical elements of the graph structure and its transformations, thereby assessing the model's comprehension of how the changes in the graph influence the classification outcome for the target node. This framework allows us to objectively measure the performance of the language model in translating structural and attribute-based differences between $G$ and $G'$ into precise and meaningful natural language descriptions.
\paragraph{Target Node Identification (TNI)} This metric assesses the ability of the LLM to accurately identify and reference the target node within the graph structure. Given the importance of the target node as the focal point of the classification task, correctly pinpointing it is crucial for generating valid explanations. Formally, given a graph $G(V, E)$, the target node $t \in V$, and a node $v \in V$ predicted by the LLM, we define the Target Node Identification (TNI) metric as:

\[
\text{TNI}(v) = 
\begin{cases}
    1 & \text{if } v = t, \\
    0 & \text{otherwise}.
\end{cases}
\]

\paragraph{Counterfactual Class Identification (CCI)} This metric evaluates the capacity of the LLM to correctly comprehend and express the change in class assignment of the target node from the factual graph $G$ to the counterfactual graph $G'$. Let $G' = (V, E')$ be the factual, such that $\model(G') = c'$ where $c'$ is the counterfactual class of the target node $t$ such that $\model(G') \neq \model(G)$, and let $c_{\text{LLM}}$ be the counterfactual class predicted by the LLM, CCI is defined as:
\[
\text{CCI}(c_{\text{LLM}}) = 
\begin{cases}
    1 & \text{if } c_{\text{LLM}} = c', \\
    0 & \text{otherwise}.
\end{cases}
\]
\paragraph{Factual Target Node Feature  (FTNF)} This metric examines the LLM's ability to accurately recognize and describe the features associated with the target node in the factual graph $G$. Correctly identifying these features is essential, as they are pivotal in determining the node's initial classification. Let $G = (V, E)$ be the factual graphs. Let $t \in V$ be the target node, and $\mathbf{x}_{t} \in \mathbb{R}^d$ denote the feature vectors of node $t$ in the factual graph $G$. We also define the vector of features predicted by the LLM for the target node $t$ as $\mathbf{x}_{\textbf{LLM}}$, the metric is computed as:
\[
\text{FTNF}(\mathbf{x}_{\textbf{LLM}}) = 
\begin{cases}
    1 & \text{if } \mathbf{x}_{\textbf{LLM}} = \mathbf{x}_{t} , \\
    0 & \text{otherwise}.
\end{cases}
\]

\paragraph{Counterfactual Target Node Feature (CFTNF)} Similar to FTNF, this metric evaluates the LLM's capability to identify and articulate the features associated with the target node in the counterfactual graph $G'$. This metric is critical for assessing whether the LLM captures the differences in node attributes that lead to a shift in classification. Let $G' = (V', E')$ be the counterfactual graphs. Let $t \in V'$ be the target node, and $\mathbf{x}'_{t} \in \mathbb{R}^d$ denote the feature vectors of node $t$ in the counterfactual graph $G'$. We also define the vector of features predicted by the LLM for the target node $t$ as $\mathbf{x}_{\textbf{LLM}}$, the metric is computed as:
\[
\text{CFTNF}(\mathbf{x}_{\textbf{LLM}}) = 
\begin{cases}
    1 & \text{if } \mathbf{x}_{\textbf{LLM}} = \mathbf{x}'_{t} , \\
    0 & \text{otherwise}.
\end{cases}
\]

\paragraph{Counterfactual Target Node Neighbors (CFTNN)} This metric measures the LLM’s capacity to correctly identify the set of neighboring nodes for the target node in the counterfactual graph $G'$. Understanding these neighbors in the counterfactual context is essential, as changes in the neighborhood structure may directly influence the target node's classification shift. Accurately capturing the neighbors in $G'$ helps the LLM articulate how the local connectivity of the target node has been modified and how this alteration affects the node’s role and classification within the network. Let $G'(V', E')$ be a counterfactual graph, let $t' \in V'$ be the target node in the graph $G'$, and let $\mathcal{N}(t') = { u \in V' \mid (t', u) \in E' }$ denote the set of neighbors of the target node $t$. Given a set of neighbors $\mathcal{N}_{\text{LLM}}(t')$ predicted by the LLM for the target node $t'$, we define the Counterfactual Target Node Neighbors (CFTNN) metric as:
\[
\text{CFTNN}(\mathcal{N}_{\text{LLM}}(t')) = 
\begin{cases}
    1 & \text{if } \mathcal{N}_{\text{LLM}}(t') = \mathcal{N}(t'), \\
    0 & \text{otherwise}.
\end{cases}
\]

Overall, these metrics provide a comprehensive framework for evaluating the language model’s capacity to interpret and describe the structural and feature-based transformations between factual and counterfactual graphs. By assessing these distinct aspects of graph understanding, we can quantitatively measure the model’s ability to generate coherent and insightful explanations that accurately reflect the graph dynamics and their implications on classification outcomes.
We decided to select the explanations that score 1 on at least 5 out of 6 metrics and test them to human judgment.

\section{Experiments}\label{sec:experiments}
Given the general framework we described in Section \ref{sec:method}, this section delves into the problem of counterfactual translation into natural language using LLM in graph neural networks. 
\subsection{Graph Counterfactual Explainers}
\begin{figure}
    \centering
    \includegraphics[width=0.5\linewidth]{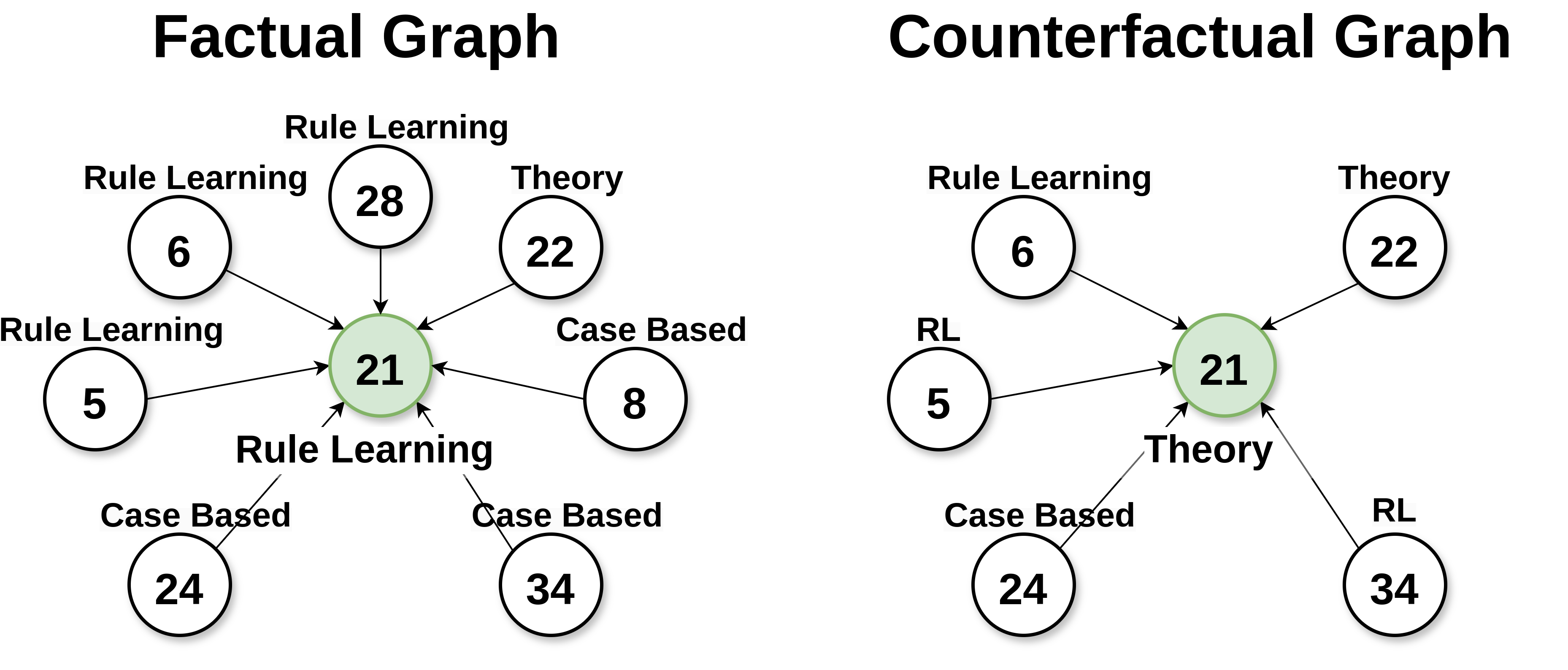}
    \caption{On the left, the factual graph. On the right, the counterfactual graph computed using CF-GNNExplainer. Each graph contains nodes ids and classes }
    \label{fig:factual-cf}
\end{figure}

As outlined in Section \ref{sec:method}, our proposed pipeline requires the integration of a counterfactual explainer for node classification, denoted as $g$. In this study, we employed two state-of-the-art explainers, namely, CF-GNNExplainer and CF-GNNFeatures, allowing us to investigate the impact of distinct graph modifications on the LLM’s ability to generate natural language explanations. Specifically, CF-GNNExplainer modifies the graph structure by perturbing the adjacency matrix, while CF-GNNFeatures focuses on altering the node attributes. This dual approach enables us to assess how variations in both structural and feature-based properties influence the quality and coherence of the generated explanations.
\paragraph{CF-GNNExplainer} is a perturbation-based counterfactual explainer. It defines $\bar{\mathbf{A}}_{v} = \mathbf{P} \odot \mathbf{A}_v$, where $\mathbf{P}$ is a binary perturbation matrix that sparsifies $\mathbf{A}_v$. The goal is to find $\mathbf{P}$ for a given node $v$ such that $\model(\mathbf{A}_v, x) \neq \model(\mathbf{P} \odot \mathbf{A}_v, x)$. To find $\mathbf{P}$, CF-GNNExplainer exploits a technique to train sparse neural networks to zero out entries in the adjacency matrix (i.e., removing edges). This results in the deletion of the edge between node $i$ and node $j$.
\paragraph{CF-GNNFeatures} is a node features perturbation-based counterfactual explainer. Given a graph $G(V, E)$, two matrices are defined, namely: $\mathbf{V}_x$, the node features matrix representing the features for every node in $G$ and  $\mathbf{P}_x$, the feature perturbation matrix. Initially, the matrix $\mathbf{P}_x$ is filled with ones to maintain the original sets of attributes. Given an oracle $\model$, parameterized by $\theta$, we fix all the weights and train $\mathbf{P}_x$ to change the attribute matrix $\mathbf{V}_x$ that is fed into the oracle multiplying the current feature matrix with the perturbation matrix. 
In Figure \ref{fig:factual-cf} you can see an example of a factual and its counterfactual graph computed using CF-GNNExplainer.
%\gabri{La figura si legge maluccio...}
\subsection{Large Language Models}

In this study, we utilized a family of state-of-the-art LLMs, namely Qwen2.5 \citep{qwen2.5}, to generate the natural language explanation. We experiment using different model sizes in terms of parameters; in particular we employ the 0.5B, 1.5B, 3B, 7B, and 14B variants.

The model configuration used in this work includes several hyperparameters that influence its behavior during text generation: 
\begin{itemize}
    \item temperature: 0.1,
    \item top-p: 0.8,
    \item top-k: 30,
    \item repetition penalty: 1.05,
    \item max output tokens: 2048,
    \item top-k: 10.
\end{itemize}
In order to reduce the memory footprint of the models, we used GPTQ \citep{frantar2023gptqaccurateposttrainingquantization} quantization with 4-bit integer precision (\texttt{int4}). 

Overall, the chosen configuration allows the model to generate meaningful counterfactual explanations for complex graph structures while maintaining computational efficiency. The setup leverages the power of modern LLMs along with advanced quantization techniques to produce high-quality outputs, making it an ideal choice for applications requiring detailed textual descriptions of graph perturbations.

\begin{figure}[h]
    \centering
    \includegraphics[width=0.5\linewidth]{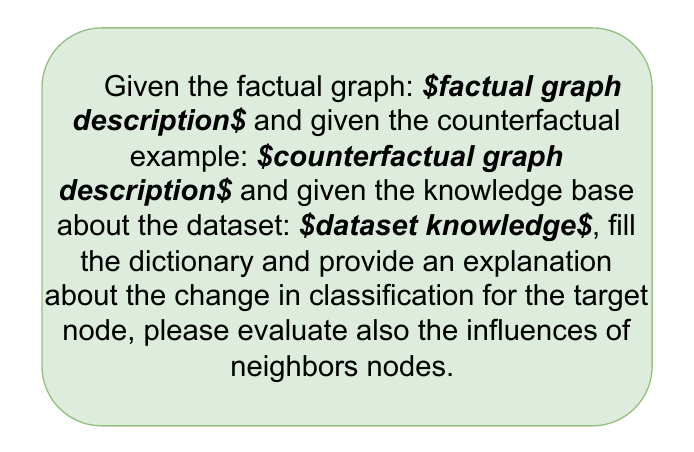}
    \caption{Prompt example to get the explanations}
    \label{fig:prompt}
\end{figure}

\paragraph{Prompting} The design and structure of the prompt is critical, as the performances of LLMs are highly influenced by how the graph data is presented. To this end, we adopted the incident representation framework proposed by \cite{fatemi2024talk}, with some adjustments to better suit the requirements of our task. To provide all the information needed for the LLM, we use an initial system prompt providing the essential background information on the challenges of counterfactual explainability, particularly in the context of graph-based data. Subsequently, the counterfactual prompt is introduced (Figure \ref{fig:prompt}), which instructs the LLM to generate a coherent and contextually appropriate natural language explanation based on both the factual and counterfactual graph samples.

\subsection{Experimental Setup}
The experiments have been carried out on a machine equipped with 64GB of RAM, an Nvidia RTX 4090, and a AMD Ryzen 9 7900 processor. As oracle we used a 2-layer GCN trained for 500 epochs with a learning rate of 0.001.

\begin{table*}[h]
\centering
\begin{tabular}{c c c c c c c c}
\toprule
\textbf{\# Parameters}& \textbf{TNI}  $\uparrow$  & \textbf{CCI}  $\uparrow$ & \textbf{FTNF}  $\uparrow$ & \textbf{CFTNF}  $\uparrow$ & \textbf{FTNN} $ \uparrow$ & \textbf{CFTNN} $\uparrow$ \\
\hline \hline
0.5B & 0.000  & 0.000 & 0.000 & 0.000 & 0.000 & 0.000 \\ 
1.5B & 0.048  & 0.048 & 0.022 & 0.007 & 0.030 & 0.030 \\ 
3B & 0.391  & 0.336 & 0.092 & 0.074 & 0.221 & 0.258 \\ 
7B & 0.292 & 0.284 & 0.162 & 0.085 & 0.007 & 0.007 \\ 
\textbf{14B }& \textbf{0.720} & \textbf{0.720} &\textbf{0.668} & \textbf{0.565} & \textbf{0.528} &\textbf{0.524} \\ 
\bottomrule

\end{tabular}
\caption{Evaluation metrics using CF-GNNFeatures and Cora dataset for graph understanding.}
\label{tab:cora}

\end{table*}

\begin{table*}[h]
\centering

\begin{tabular}{c c c c c c c c}
\toprule
\textbf{\# Parameters} & \textbf{TNI}  $\uparrow$  & \textbf{CCI}  $\uparrow$ & \textbf{FTNF}  $\uparrow$ & \textbf{CFTNF}  $\uparrow$ & \textbf{FTNN} $ \uparrow$ & \textbf{CFTNN} $\uparrow$ \\
\hline \hline
0.5B & 0.000  & 0.000 & 0.000 & 0.000 & 0.000 & 0.000 \\ 
1.5B & 0.053  & 0.053 & 0.039 & 0.026 & 0.000 & 0.013 \\ 
3B & 0.237 & 0.197 & 0.105 & 0.132 & 0.066 & 0.105 \\ 
7B & 0.171  & 0.118 & 0.053 & 0.118 & 0.026 & 0.000 \\ 
\textbf{14B} & \textbf{0.618}  & \textbf{0.618} & \textbf{0.579} & \textbf{0.579} & \textbf{0.539} & \textbf{0.500} \\ 
\bottomrule

\end{tabular}
\caption{Evaluation metrics using CF-GNNExplainer and Cora dataset for graph understanding.}
\label{tab:coracfgnn}
\end{table*}

\begin{table*}[h]
\centering

\begin{tabular}{c c c c c c c c}
\toprule
\textbf{\# Parameters} & \textbf{TNI}  $\uparrow$  & \textbf{CCI}  $\uparrow$ & \textbf{FTNF}  $\uparrow$ & \textbf{CFTNF}  $\uparrow$ & \textbf{FTNN} $ \uparrow$ & \textbf{CFTNN} $\uparrow$ \\
\hline \hline
0.5B & 0.000  & 0.000 & 0.000 & 0.000 & 0.000 & 0.000 \\ 
1.5B & 0.273  & 0.280 & 0.096 & 0.003 & 0.174 & 0.174 \\ 
3B & 0.488 & 0.394 & 0.112 & 0.037 & 0.320 & 0.348 \\ 
7B & 0.450  & 0.447 & 0.180 & 0.171 & 0.314 & 0.314 \\ 
\textbf{14B} & \textbf{0.879} & \textbf{0.873} & \textbf{0.705} & \textbf{0.481} & \textbf{0.758} & \textbf{0.758} \\ 
\bottomrule

\end{tabular}
\caption{Evaluation metrics using CF-GNNFeatures and Citeseer dataset for graph understanding}
\label{tab:citeseer}
\end{table*}

\begin{table*}[h]
\centering

\begin{tabular}{c c c c c c c c}
\toprule
\textbf{\# Parameters}  & \textbf{TNI}  $\uparrow$  & \textbf{CCI}  $\uparrow$ & \textbf{FTNF}  $\uparrow$ & \textbf{CFTNF}  $\uparrow$ & \textbf{FTNN} $ \uparrow$ & \textbf{CFTNN} $\uparrow$ \\
\hline \hline
0.5B & 0.000  & 0.000 & 0.000 & 0.000 & 0.000 & 0.000 \\ 
1.5B & 0.000  & 0.038 & 0.000 & 0.000 & 0.000 & 0.000 \\ 
3B & 0.385 & 0.192 & 0.038 & 0.077 & 0.115 & 0.115 \\ 
7B & 0.385  & 0.346 & 0.154 & 0.269 & 0.077 & 0.038 \\ 
\textbf{14B} & \textbf{0.615}  & \textbf{0.615} & \textbf{0.346} & \textbf{0.346} & \textbf{0.577} & \textbf{0.538} \\ 

\bottomrule

\end{tabular}
\caption{Evaluation metrics using CF-GNNExplainer and Citeseer dataset for graph understanding.}
\label{tab:citeseercfgnn}
\end{table*}

\subsection{Datasets} To ensure a comprehensive evaluation, we employed two well-known citation network datasets: Cora and CiteSeer. The CiteSeer dataset comprises 3,312 scientific publications categorized into six distinct classes, connected by 4,732 citation links. Each publication is represented by a binary word vector, where each entry indicates the absence or presence of a specific word from a dictionary of 3,703 unique terms. Similarly, the Cora dataset contains 2,708 scientific publications classified into seven distinct classes, with 5,429 citation links. Each publication is also represented by a binary word vector corresponding to a dictionary of 1,433 unique terms.

After the counterfactuals had been found, we translated the original node feature vocabulary using actual words rather than binary vectors to facilitate the language model’s understanding of the relationship between a node’s classification and its features. Since predefined vocabularies for node features are not provided for these datasets, we adhered to the original feature extraction instructions to generate the vocabularies for both datasets.

\subsection{Results} To provide an appropriate evaluation, we tested multiple counterfactual methods and LLMs on different graph datasets. As shown in Tables from \ref{tab:cora} to \ref{tab:citeseercfgnn} the results for the graph understanding as defined in Section \ref{sec:method} are generally good for models with more parameters. In particular, a clear trend emerges across all tables: as the number of parameters in the LLM increases, performance improves significantly across all metrics, as expected. The smallest model, with 0.5 billion parameters, consistently scores zero across all metrics and datasets, indicating its inability to generate meaningful explanations. With an increase to 1.5 billion parameters, there is a minimal improvement, but performance remains substantially low.

Notable improvements are observed with the 3-billion-parameter model, especially in metrics like Target Node Identification (TNI) and Counterfactual Class Identification (CCI). However, it's the largest model, with 14 billion parameters, that achieves the highest scores across all metrics and datasets. This demonstrates the importance of model size when dealing with complex tasks such as generating natural language explanations from graph counterfactuals.

In Table \ref{tab:cora}, results show that using CF-GNNFeatures on the Cora dataset, the TNI metric increases from 0.000 with the smallest model to 0.720 with the largest model. Similarly, in Table \ref{tab:citeseer}, using the CiteSeer dataset, the TNI metric reaches 0.879 with the 14-billion-parameter model. These improvements indicate that larger models can understand and generate accurate descriptions of complex graph structures.

\paragraph{Comparing performances across datasets} The LLMs generally perform better on the CiteSeer dataset, especially when using the CF-GNNFeatures explainer. For example, in Table \ref{tab:citeseer} (CF-GNNFeatures on CiteSeer), the 14-billion-parameter model achieves a TNI of 0.879, higher than the 0.720 achieved on the Cora dataset in Table \ref{tab:cora}. 

Several factors contribute to this difference in performance. The datasets have different levels of complexity, differences in feature distributions, or inherent properties that make one more amenable to the LLM's processing capabilities. CiteSeer have more straightforward or more distinctive features that the LLM can more readily associate with classification outcomes, aiding in generating coherent explanations.

\paragraph{Comparison between explainers} The two counterfactual explainers used in the study, CF-GNNFeatures and CF-GNNExplainer, focus on different aspects of the graph. CF-GNNFeatures modifies node features, while CF-GNNExplainer modifies the graph structure.

For instance, in Table \ref{tab:cora} (CF-GNNFeatures on Cora), the FTNF (Factual Target Node Feature) and CFTNF (Counterfactual Target Node Feature) scores at 14 billion parameters are 0.668 and 0.565, respectively. In contrast, in Table \ref{tab:coracfgnn} (CF-GNNExplainer on Cora), these scores are lower, at 0.579 for both metrics at the same model size.

This suggests that LLMs find it easier to generate explanations when the modifications involve changes in node features rather than structural changes in the graph. Explaining structural changes may require a deeper understanding of the graph's topology and how it influences the classification, which appears more challenging for the LLMs.

\begin{table}[htbp]
\centering
\begin{tabular}{c|c|c}

\hline
\textbf{Question} & \textbf{CF-GNNE.} & \textbf{CF-GNNF.} \\
\hline 
Q1 & 3.00 $\pm$ 1.60 & 4.44 $\pm$ 0.80 \\
Q2 & 2.87 $\pm$ 1.47 & 4.28 $\pm$ 0.83 \\
Q3 & 2.90 $\pm$ 1.51 & 3.56 $\pm$ 1.23 \\
Q4 & 2.92 $\pm$ 1.52 & 3.68 $\pm$ 1.20 \\
Q5 & 3.00 $\pm$ 1.55 & 3.80 $\pm$ 1.11 \\
\hline
\end{tabular}
\caption{Average scores (1 to 5) from the human evaluation for the explanations generated using Qwen2.5-14B using counterfactuals from CF-GNNExplainer (CF-GNNE.) and CF-GNNFeatures (CF-GNNF) on the Cora dataset}
\label{tab:humaneval}

\end{table}

\paragraph{Human Evaluation}
In order to assess appropriately the explanations, we built a questionnaire asking to answer the following questions using a scale from 1 to 5: 
\begin{itemize}
    \item Q1: Is the terminology and language used in the explanation appropriate and easy to understand?
    \item Q2: How clear and easy to understand is the provided counterfactual explanation?
    \item Q3: How clearly does the explanation describe the changes in node connections (graph structure) that led to the counterfactual outcome?
    \item Q4: Are the changes in features and structure easy to interpret and make sense in the context of the original graph?
    \item Q5: What is your overall assessment of the clarity and coherence of the counterfactual explanation?
\end{itemize}
The results of the human evaluation can be seen in Table~\ref{tab:humaneval}. The human evaluation shows that explanations generated by the LLM that focus on node features are more effective and meaningful to human users than those focusing on adjacency matrix perturbations: participants found feature-based explanations to be more accessible in terms of language, clearer in conveying the reasoning, and more interpretable within the context of the graph. 

\section{Practical Implications}
\label{sec:discussion}
The proposed framework for generating natural language counterfactual explanations using LLMs holds significant promise for enhancing interpretability and transparency in graph-based machine learning models. This approach has several practical implications across domains where complex graph structures are used, such as financial analysis, healthcare, cybersecurity, and social network analysis.
By translating complex counterfactual explanations into natural language, our method makes these explanations more accessible to non-technical stakeholders, including business managers, policymakers, and end-users. Moreover, with the European Union’s General Data Protection Regulation (GDPR) and the EU Artificial Intelligence Act, explainability and transparency of AI systems are becoming mandatory, particularly in high-stakes applications. Our method addresses this requirement by ensuring that counterfactual explanations are technically sound and easily interpretable, thereby aiding compliance with legal and ethical standards.
In summary, our framework provides a comprehensive solution for bridging the gap between technical counterfactual explanations and human comprehension, enabling broader adoption of AI technologies in high-stakes, complex domains. The method promotes transparency and trust and enhances the utility of counterfactual explanations as a tool for model evaluation, debugging, and refinement.

\section{Conclusion and Future Work}
\label{sec:conclusion}
In this paper, we presented a novel framework for generating natural language counterfactual explanations for graph-based models using state-of-the-art LLMs. Our approach leverages the inherent generative capabilities of LLMs to transform complex and technical counterfactual examples into coherent and accessible natural language descriptions. By doing so, we address a critical gap in the literature: the lack of intuitive, user-friendly counterfactual explanations for GNNs. We validated our method across several GNN-based counterfactual explainers and multiple graph datasets, demonstrating its effectiveness through a suite of newly proposed metrics and human evaluation. Our findings show that our framework not only captures the underlying structural and attribute-based transformations within the graphs but also produces explanations that are easily interpretable by non-expert users, thereby enhancing the transparency and usability of graph-based machine learning models.

Our framework does not depend upon any particular counterfactual explainer, so it can be used with any approach. We developed additional metrics that quantify the quality of the natural language output—such as graph understanding. Finally, our experiments were conducted using open-source LLMs without task-specific fine-tuning. Future research could explore the impact of fine-tuning the LLMs on graph-specific tasks or developing domain-specific LLMs to further enhance the quality and relevance of the generated explanations.

Overall, we believe that our framework represents a significant step toward bridging the gap between complex algorithmic explanations and human comprehension in the domain of graph-based machine learning. 
%Bibliography
\bibliographystyle{plainnat}  
\bibliography{references}

\end{document}